# Deciding Without Pointing: A Case Study of Whether a Decision Model Localizes the Anomalies It Detects

Hsiang-Hung Lee
Politecnico di Milano
Milan, Italy
hsianghung.lee@mail.polimi.it

## Abstract

*Decision models can answer a typed question about an image with a probability per option, without generating text. In our case study of Clef-flash (post-trained Qwen3.5-9B backbone) on industrial inspection benchmarks, we ask whether it localizes the anomalies it detects. We obtained uneven results with zero-shot detection: per-category image AUROC 0.924 on MVTec AD and 0.668 on MVTec AD 2. On the former, the decision head's routing attention produces 0.391 mean per-image pixel AUROC, and hiding the image tokens from it leaves detection unchanged. The multimodal LLM's backbone attention does point at the defect, however it falls below a center prior when we change question. It falls below chance if we remove textual context in the question, partly due to backbone attention leaving the object, while the head's map hardly changes. The backbone's states carry the location when probed with normal images, on the test-split images, however, the decision depends on the defect region. A larger model, Clef, that comes with a post-trained Qwen3.8-27B backbone, gives no useful map in the decision head either. We found that it's possible to recover anomaly location by asking question specifically on an image region, retraining the head or reading the backbone, but none matches dedicated few-shot localizers.*

## 1. Introduction

An automated visual inspection station in factories has to decide whether a part is defective. When it flags one, it should show an operator where to look. Zero-shot methods based on vision-language models promise to do both without prior training on the images of the product [1], [2]. A new class of model, the *decision model*, answers a typed question about an image ("does this part have a defect? true or false") with a probability per option, computed by a small head over a large multimodal backbone, without generating text [3], [4]. Its head attends over all input positions, so its attention looks like a map of where the model looked, and reading attention as a map is common practice for generative multimodal models [5], [6]. Clef and Clef-flash don't return attention weights, but an explanation tool can recompute them with a hook, as we do. The one evaluation of these models on industrial images that we found, which is not peer reviewed, only compares detection accuracy [7], and we found no studies on the localization capabilities of this kind of models.

We present a case study of one released decision model, Clef-flash, on four public benchmarks. The study asks four questions: does it detect zero-shot, and is its probability calibrated; does the head's attention localize the defect; is the information present and does the decision depend on the defect region; and how can a location be recovered, at what cost. Plausibility (agreement with the masks) and faithfulness (whether a map reflects what the decision depends on) are kept apart throughout the study [8]. Our contributions are:

- readouts that separate decision, head attention, backbone attention, information content and dependence (R0–R7, Table 1);
- evidence of a dissociation in a decision model: on MVTec AD the head's attention ranks the defect low (per-image pixel AUROC) while the information is present, the backbone's attention beats a center prior only when explicitly given an anomaly inspection question, and, on the test-split images, the decision depends on the defect region; across both releases, the head's attention is not a useful map;

- a characterization of zero-shot detection against baselines and against the model's own backbone answer (prior exposure unknown, Sec. 5), and of its calibration;
- three tested methods to recovering location (asking, teaching, reading) with their costs, compared with zero-shot and few-shot localizers on the same images.

All experiments were preregistered and controlled to prevent HARKing (Hypothesizing After Results are Known) or P-hacking.

## 2. Related work

**Industrial anomaly detection.** CLIP-based methods score image patches against text prompts for normal and anomalous states [1], [2], [9]; memory-bank methods compare patches with normal images [10].

**Attention of multimodal models as a map.** ReAL reads the attention of a generative model over image tokens as an anomaly map [5]; EAGLE reports that correct predictions go with more attention on the defect [6]; IADGPT trains such a map with masks [11]. These maps are read from generation; a decision head is a different reader. Benchmarks of multimodal language models on industrial inspection score their generated answers [12].

**Attention as explanation.** Whether attention explains a decision is contested [13], [14]; we follow the distinction between plausible and faithful explanations [8], and test dependence by occluding image regions [15], in the spirit of removal-based evaluation [16], though without retraining and with blocks of normal images of the same product. AnomalyJev evaluates text explanations against ground truth and calls that score faithfulness [17], which we call plausibility; we adopt its 3 × 3 grid rule for region claims.

**Logical anomalies.** LogicAD and LogSAD use text about normal composition on MVTec LOCO AD, with reference images and separate reasoning or matching steps [18], [19]. Our test gives one simple prompt to a model that sees no normal image.

## 3. Model and readouts

Clef-flash combines a post-trained Qwen3.5-9B multimodal backbone with a joint schema head [4]. The head was trained jointly with rank-256 low-rank adapters [3], and the release ships a merged backbone. A partial comparison with the base model's weights finds differences only in the language model's linear projections, so "the backbone" here is the post-trained one. A request carries a state text, the product name, a note and a question. The image becomes a 16 × 16 grid of merged-patch tokens at $512^2$. The head projects all backbone states into a memory, routes each option over that memory in two evidence-routing layers, and scores each option from summary vectors of the question, the options and the last token, plus a gated contribution from its attention over the memory.

Figure 1 shows where each readout is taken and Table 1 lists them. R1 is the head-mean over the two routing layers and their 16 heads of the attention that the option "true" pays to the image positions, laid out on the 16 × 16 grid in raster order and not renormalized; R1-dec is the question field's cross-attention in the four decoder layers, averaged likewise. R2 averages the attention from the question tokens to the image tokens over the eight full-attention layers of the backbone, which instead uses linear attention. R5, a gradient readout, was not run due to hardware constraints. The models return no attention weights, so we recompute the head's attention with a forward hook in fp32, check it against the module's own output, and report head-averaged maps only since single heads are numerically fragile across devices.

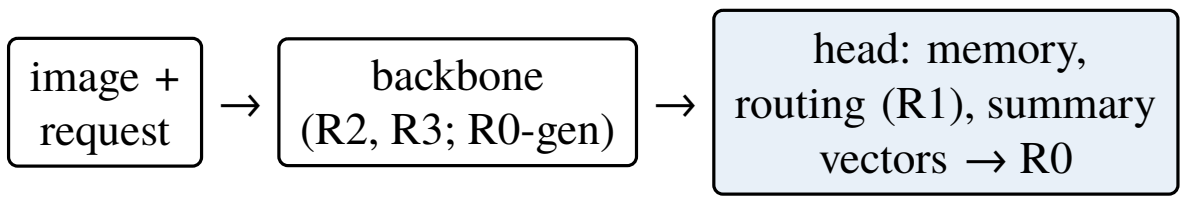

R4: replace a block of the image; R6: add region questions; R7: retrain the head.

Figure 1. Schematic: where the readouts are taken.

| Readout | What it measures |
|---|---|
| R0 | the decision: the probability of "true" for the anomaly question |
| R0-gen | the backbone's own yes/no answer from the same weights, without the head |
| R1 | the head's routing attention over image tokens (R1-dec: its decoder cross-attention) |
| R2 | the backbone's attention from the question tokens to the image tokens |
| R3 | a nearest-neighbor probe of the states against 60 normal images (information, not explanation) |
| R4 | the drop of the decision margin when one block of an 8 × 8 grid is replaced by the matching block of a normal image (dependence) |
| R6 | one additional yes/no question for each cell of a 3 × 3 grid, in the same pass (asking) |
| R7 | the routing attention of a head retrained on synthetic defects (teaching) |

Table 1. Readouts. All maps are post-processed similarly and scored by per-image pixel AUROC on defective images.

## 4. Protocol

Every experiment was pre-registered with its hypotheses, controls and decision rules. For the datasets, we use MVTec AD [20], VisA [21], MVTec AD 2 [22] and MVTec LOCO

AD [23] (image level only). The dev split is 30% of each benchmark's test set per category and label. Choices of layer, phrasing and threshold were made on the same split. The teaching experiment chose its loss weight and stopping on a validation split of its source dataset, so the target is never seen. Numbers are on the full test set (dev images included) for configurations fixed in advance and on the 70% test split for selected ones. The two follow-ups cover MVTec AD only. The one with changed requests uses the test split only, for cost, a deviation from the rule above.

The main metric is the per-image pixel AUROC on defective images. Pooled pixel AUROC and AUPRO (to 30% false-positive rate) are reported for reference. Every map is upsampled bilinearly to $256^2$ and smoothed with a Gaussian of $\sigma = 4$ px; masks are resized to $256^2$ by nearest neighbor. Maps share one post-processing and are compared with a center prior, a same-category shuffle (another image's map of the same product), a random map and a foreground prior. The center prior is an isotropic Gaussian at the image center on the token grid. Differences use paired tests with Holm correction and 95% bootstrap intervals over images (2,000 resamples). We say that a map ranks the defect low when its mean per-image pixel AUROC is below 0.5. This does not claim that the attention responds to the defect, which was not tested.

The decision margin is the logit of "true" minus that of "false", and the default threshold is margin 0. Direct comparisons of image AUROC, unless another rule is stated, use four labels with a ±0.03 margin. A comparison is equivalent if the 95% interval of the difference lies inside ±0.03, better or worse if it lies wholly beyond the margin, and inconclusive otherwise. An interval that excludes zero but crosses the margin is therefore inconclusive.

Our zero-shot WinCLIP run does not reproduce its published numbers on MVTec AD and is labeled as our run. In percent, image AUROC is 90.3 against 91.8 published [1] (−1.506 points, just outside the ±1.5-point tolerance). Pixel AUROC is 82.4 against 85.1, and PRO is 62.1 against 64.6. The PRO values are reported only (its native pipeline); Table 13 uses our shared post-processing. About one point of the image-level gap matches a multi-crop step that our implementation lacks; the rest is not explained. Each gap lowers the baseline, so direct MVTec AD comparisons with this run lean toward Clef-flash. The published image AUROC lies inside the interval of the decision's.

The model runs in bf16 on a NVIDIA GeForce RTX 5080 that comes with 16 GB VRAM, with part of the backbone offloaded to the CPU (Intel® Core™ 14900K). The models were never quantized.

| **Quantity** | **Value [95% CI]** |
|---|---|
| *Image AUROC of the decision, per-category mean* | |
| MVTec AD | 0.924 [0.910, 0.937] |
| VisA | 0.847 [0.831, 0.862] |
| MVTec AD 2 | 0.668 [0.613, 0.725] |
| MVTec LOCO AD, structural | 0.710 [0.679, 0.742] |
| MVTec LOCO AD, logical | 0.466 [0.432, 0.499] |
| MVTec LOCO AD, logical, with the composition sentence | 0.705 [0.674, 0.735] |
| *Differences, with their reading* | |
| Decision − our WinCLIP, MVTec AD: inconclusive | 0.021 [0.000, 0.041] |
| The same, decision at $256^2$ (exploratory) | −0.027 [−0.050, −0.004] |
| Decision − R0-gen, MVTec AD: equivalent | 0.006 [0.000, 0.013] |
| Decision − R0-gen, VisA: equivalent | 0.000 [−0.008, 0.009] |
| Change in the advantage over our WinCLIP, MVTec AD 2: no evidence of contamination | 0.099 [0.024, 0.169] |
| Drop under perturbations, decision − our WinCLIP: inside ±0.02 | 0.000 [−0.006, 0.007] |

Table 2. Zero-shot detection. Readings use the rules of Sec. 4 and, for the last two rows, the pre-registered rules in the text.

## 5. Zero-shot detection

*Zero-shot* here means that the model sees no image of the product prior detection and is not adapted to it. Of the product, the request carries only its name. The decision detects unevenly (Table 2). On MVTec AD pill is close to chance (0.517), and on MVTec AD 2 five of eight products are not shown above chance (exploratory per-product intervals). Against our zero-shot WinCLIP run the comparison is inconclusive. WinCLIP sees $240^2$ inputs against $512^2$, so Table 2 also gives the difference with the decision at $256^2$ (exploratory).

The backbone's own answer ranks images as well as the head on MVTec AD and VisA, both equivalent within ±0.03 (Table 2). The shipped backbone's own answer reaches the same discrimination, so this test sees nothing the head module adds.

The vendor's training data are not public, so the benchmarks may have been seen. Neither the fall from MVTec AD to MVTec AD 2 nor the two results below can exclude that. We report MVTec AD 2 as a generalization result on a newer benchmark.

A model that had seen MVTec AD might lose its advantage over a baseline on newer data. On MVTec AD 2 (regular lighting, public test images plus validation normals, 486 images) Table 2 gives the change in the advantage over our WinCLIP run. The one-sided rule (lower end above −0.05) was pre-registered after the perturbation results. By this rule, the change means no evidence of contamination (prior exposure to MVTec AD). The validation images narrowly passed the pre-registered exchangeability check. The analyses without them and across all lighting conditions give the same outcome.

The test is weak. WinCLIP is close to chance there (0.548), so the statistic mostly reflects whether the decision stays above chance. Three things can push it upward: the resolution gap, possible exposure to MVTec AD 2 (released with a 2025 preprint [24]), and WinCLIP's own evaluation on MVTec AD in its paper.

Robustness to label-preserving perturbations is equivalent to WinCLIP's. We applied six of them (rotation by ±5° with about 9% zoom, a 95% crop, JPEG at quality 90, brightness ±10%) to 13 of the 15 categories. Transistor and metal nut are excluded because their defects depend on orientation. Over 1,510 images, the difference between the decision's drop and WinCLIP's lies inside the pre-registered ±0.02 band (Table 2).

On MVTec LOCO AD the decision ranks structural anomalies above good images and is not shown to detect logical ones (Table 2). One sentence stating the normal composition, fixed before any image was seen, raises the logical AUROC (the same per-product mean). At the default threshold, however, the model then flags good parts almost as often as anomalous ones.

On MVTec AD, at the benchmark's defect rate, the probability is badly calibrated (ECE 0.392 [0.375, 0.409]), too low for defective images. Much of the offset depends on the answer format, in a test designed after the wording test below and run on the same test images. Attaching "defective" to the option "false" lowers the ECE on the test split from 0.390 to 0.237. The ranking is nearly unchanged (0.922 and 0.909, descriptive, no paired interval). The wording of the question moves the ranking little (spread 0.006).

A threshold from thirty normal images, which makes the operating point few-shot, raises the recall on the MVTec AD test split from 0.436 to 0.712, at a false-positive rate of 0.036.

## 6. Decision without localization

**The head's attention ranks the defect low.** Over the 1,258 defective MVTec AD test images, R1 scores below chance and below every control (Table 3, Figure 2). It also scores below another image's map of the same product (paired difference −0.170 [−0.184, −0.156]), so the low ranking is specific to each image. R1-dec, the head's decoder cross-attention, lies below chance too. A map reliably below chance carries inverted information. We do not report the inverted map as a localizer, because the sign would be chosen after seeing the result and the larger release does not share it.

| Map | MVTec AD | VisA |
|---|---|---|
| R1 | 0.391 [0.381, 0.402] | 0.443 [0.430, 0.455] |
| R1 within the object | 0.289 [0.275, 0.302] | 0.373 [0.360, 0.385] |
| R1-dec | 0.323 [0.313, 0.333] | — |
| Shuffle of R1 | 0.561 [0.551, 0.571] | 0.529 [0.516, 0.543] |
| Center prior | 0.760 [0.750, 0.769] | 0.824 [0.816, 0.833] |

Table 3. Clef-flash's head attention and two controls: per-image pixel AUROC on defective test images (means, 95% CIs). The within-object row covers the categories with a valid foreground, and VisA's value is a secondary analysis; — means not reported here.

On VisA R1 is below chance as well, against a higher score for the shuffle (Table 3). On MVTec AD the head attends more to the object than to the background, and within the object it scores lower still, so this is not a background preference. The same holds on VisA (a secondary analysis). Image tokens with outlying norms, which could act as attention sinks [25], are too rare to explain it. Hiding them changes the map by −0.002 against hiding as many random tokens (descriptive).

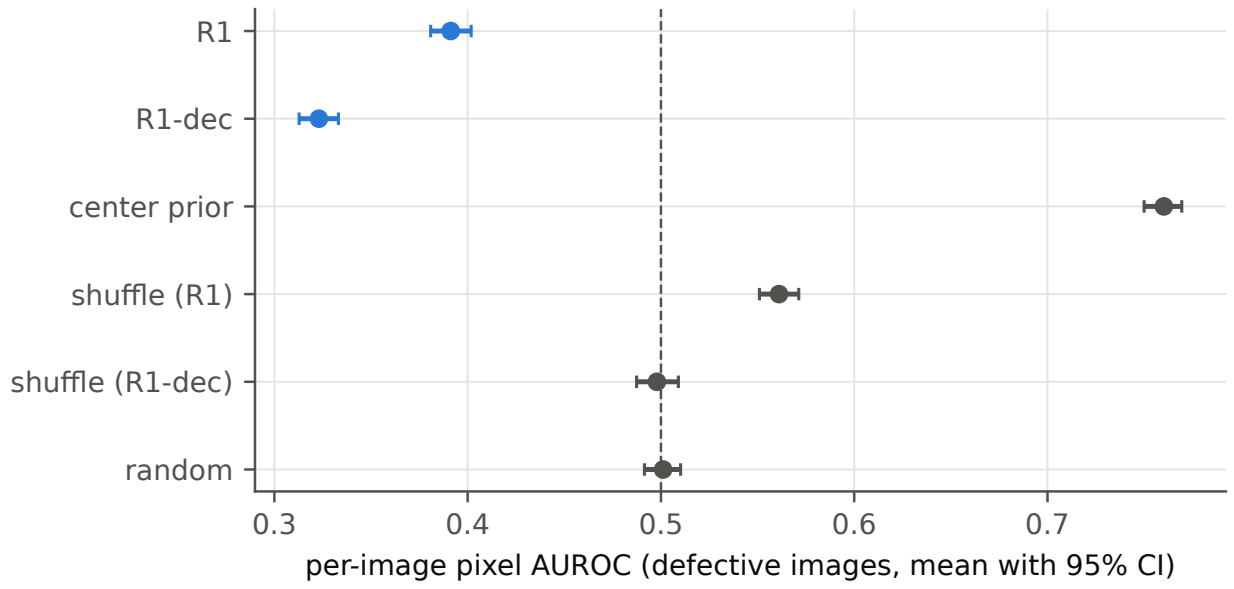

Figure 2. Per-image pixel AUROC on defective MVTec AD test images of the head's routing attention R1, its decoder cross-attention R1-dec, and the controls (means, 95% CIs). Both head readouts lie below chance and below every control.

**A follow-up qualifies the low ranking.** This analysis of the stored maps was written and pre-registered after a review of the first results. Across products the equal-category mean is 0.406 with a t interval of [0.347, 0.464]. This is a guard on the sign, not inference to a population of products.

Each category is labeled by a one-sample Wilcoxon test with Holm correction over the 15 and the bootstrap interval of its mean. The label is below or above when both agree, null when the test is not significant and the interval contains 0.5, and inconclusive when the two disagree. This is a different rule from the image-level labels of Sec. 4. Table 4 gives the counts, on the full set and, in an amendment after first results, on the test split. The low ranking therefore holds in about half of the products and not in every product. The pre-registered rule called fewer than 8 of 15 a minority of products; the full-set count sits on that line and the test-split count is just under it.

| Tested quantity (of 15 products) | Below | Above | Null | Inconcl. |
|---|---|---|---|---|
| R1 against 0.5, full set | 8 | 1 | 3 | 3 |
| R1 against 0.5, test split | 7 | 1 | 4 | 3 |
| Single-block contrast against 0, test split | 2 | 6 | 6 | 1 |

Table 4. Per-product labels on MVTec AD. The last row is the dependence contrast of Table 7; both of its products labeled below are named in the text. Post-results follow-ups.

R1 is below 0.5 on the images detected at the default threshold and, on the same categories, on those missed (Table 5; 13 categories with at least five such images, which leaves out cable and toothbrush). At a threshold from thirty normal images (Sec. 5) both are also below 0.5. Whether they differ is not shown: the stratified difference between them changes sign with the threshold.

The direction also holds on attention mass (log concentration on defect cells −0.279 [−0.300, −0.258]), a weak addition, since it is close to implied by the rank metric. On other measures R1 is worse than a control rather than below chance. Its top cell on the 16 × 16 token grid hits the defect in 7.2% of images against 10.3% for the shuffle (difference −0.030 [−0.049, −0.011]). Its average precision minus the center prior's is −0.083 [−0.092, −0.073].

**The decision barely uses this attention.** Hiding all 256 image tokens from the head's six attention modules leaves the image AUROC equivalent (difference 0.000, [−0.001, 0.001]), and no image changes its decision at the default threshold. This is consistent with the architecture. The memory term is scaled down by a learned gate, and the image reaches the decision through the summary vectors, which the backbone fills while it reads the request. By the pre-registered test the attention is faithful to its own path:

| Threshold | Detected | Missed | Stratified difference |
|---|---|---|---|
| Default | 0.365 [0.349, 0.381] | 0.376 [0.356, 0.394] | 0.014 [−0.011, 0.037] |
| 30 normal images | 0.403 [0.391, 0.416] | 0.432 [0.407, 0.456] | −0.046 [−0.076, −0.017] |

Table 5. R1′s per-image pixel AUROC on detected and missed MVTec AD images (95% CIs). The first two columns are equal-category means. The last is detected minus missed within each category, weighted by the sizes of the two groups, so it need not equal the difference of the first two. The columns cover 13, 13 and 15 categories at the default threshold and 15, 10 and 12 at the other. A post-results follow-up.

replacing its 32 most-attended positions moves the margin more than 32 random ones, by 0.005 logits. The effect, however, is negligible in size.

**The information is present.** The backbone's attention from the question tokens, averaged over its eight full-attention layers, scores 0.863 [0.854, 0.873], above every control (Figure 3). This is agreement with the masks, not evidence that the decision uses these positions.

The probe of the final states with sixty normal images reaches 0.897. That is above the center prior by 0.137 [0.125, 0.149] and above the foreground prior. On VisA it is shown within the object (exploratory) and only marginally on the whole image. The probe is an information test, not a competitive map: PatchCore with sixteen images is better than it with as many (−0.085, [−0.092, −0.078]).

On Clef-flash, EAGLE's link between attention on the defect and a correct answer agrees in its pooled direction for the backbone's own generated answer (Holm-adjusted p = 0.010, heterogeneous across products). It is not detected for the head's decision.

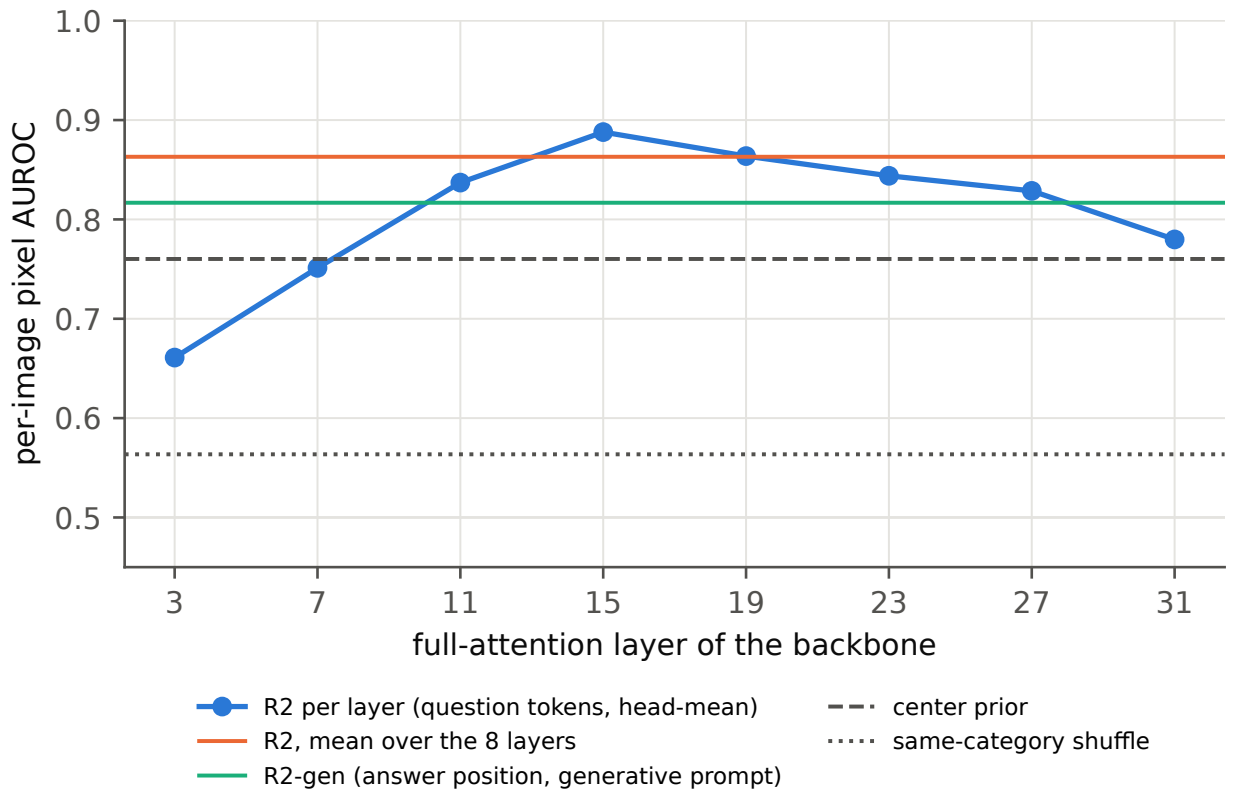

Figure 3. Per-image pixel AUROC of the backbone's attention from the question tokens to the image tokens, per full-attention layer (MVTec AD test images, head-mean within the layer), with the controls as horizontal lines. R2-gen is the same attention from the answer position of a generative prompt. Middle layers localize best.

**The backbone's attention depends on the request.** A second follow-up keeps the image and changes the request. The default inspection request (D) has the task "Industrial visual inspection", the product name and the note "The image shows one {name} from a production line.". Its field is "defective" and its question is "Does this {name} have a visible defect, damage, contamination, or any other anomaly?". W1 and W2 keep all but the sentence: "Does this image show a {name}?" and "Is this image in color?".

C1 and C2 use the same two sentences with the task "Image description", no note, no product field and the field "answer" (C1 and C3 still name the product in the sentence).

| Arm | Request | R2 | R2 − center prior | D − arm | R2 within the object |
|---|---|---|---|---|---|
| D | default | 0.862 [0.851, 0.874] | 0.103 [0.085, 0.121] | — | 0.865 [0.850, 0.879] |
| W1 | sentence changed | 0.708 [0.693, 0.724] | −0.051 [−0.073, −0.030] | 0.154 [0.145, 0.163] | 0.768 [0.750, 0.786] |
| W2 | sentence changed | 0.613 [0.598, 0.628] | −0.146 [−0.167, −0.126] | 0.249 [0.239, 0.258] | 0.688 [0.670, 0.706] |
| C1 | no context | 0.271 [0.261, 0.280] | −0.489 [−0.505, −0.473] | 0.592 [0.579, 0.605] | 0.355 [0.336, 0.373] |
| C2 | no context | 0.314 [0.304, 0.323] | −0.446 [−0.463, −0.429] | 0.549 [0.537, 0.562] | 0.437 [0.419, 0.454] |
| C3 | no context, default's length | 0.334 [0.323, 0.345] | −0.426 [−0.444, −0.407] | 0.528 [0.517, 0.540] | 0.520 [0.500, 0.540] |
| P | no context, positive control | 0.389 [0.377, 0.401] | −0.370 [−0.388, −0.351] | 0.473 [0.460, 0.485] | 0.479 [0.461, 0.497] |

Table 6. The backbone's attention under each request: per-image pixel AUROC on defective MVTec AD test-split images and paired differences (means, 95% CIs). D − C1 and D − C2 are the primary contrasts, D − W1 and D − W2 the wording contrasts; the other differences are secondary and the within-object column is exploratory (eight categories). D − P over all categories is descriptive; the registered positive control, on the eight categories with a valid foreground, is 0.509 [0.494, 0.523]. A post-results follow-up.

C3 ("Was this {name} photographed with a camera, phone, scanner, or any other device?") has the default's number of question tokens. P ("Is the background of this image uniform?") is a positive control. Both are without the context. The follow-up was written after these results, with its requests fixed before it ran. Its prediction that R2 stays above the center prior under neutral requests was refuted (Figure 4).

On the test split, "points at the defect" means above chance and above the controls (Table 6). Under D, R2 is above the center prior and the foreground prior (0.073 [0.053, 0.093]). With the sentence changed and the context kept it is still above the shuffle but below the center prior. These sentences are also shorter than the default's. The drop from D under W1 and W2 means that the sentence matters given the context.

With the context removed as well it falls below chance, and D minus C1 and D minus C2 are positive in all 15 categories. No changed request retains any of the default's excess over the center or foreground prior. The context-free arms change the sentence, task, product, note and field together, so this does not show which part of the context matters.

In the default, the rows of the anomaly words and the other rows score 0.882 and 0.852 (difference 0.030 [0.026, 0.035]), so the map is not confined to those rows. This compares rows and does not test what the words contribute.

Image tokens come before the request and the backbone is causal. Final-layer image-token states were identical across requests on 20 checked images of one category (a corrected check, re-run after first results because the registered one could not fail). The change therefore concerns where the question tokens read, not what the image-token states hold.

Part of the below-chance score is the map leaving the object. Within the object (exploratory, eight categories, Table 6) C1, C2 and P stay below 0.5 and C3 does not.

R1 stays between 0.383 and 0.391 under every request (D minus C1 0.001 [−0.002, 0.005], median map correlation with D 0.975). But the routing query barely changes (median cosine with D 0.931 for C1 and 0.955 for P), so this test cannot separate a cause in the image from one in the request.

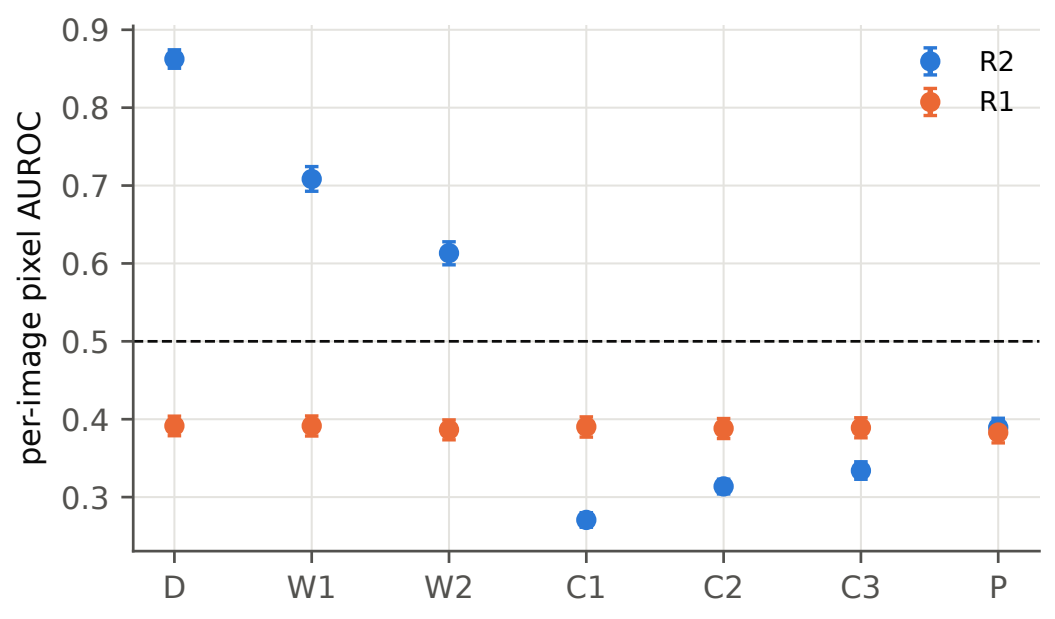

Figure 4. Per-image pixel AUROC on defective MVTec AD test-split images (means, 95% CIs) of R2 and R1 under the default request (D), with the sentence changed and the context kept (W1, W2), and with the context removed (C1, C2, C3, P); the dashed line is chance. A post-results follow-up.

**The decision depends on the defect region, on these images.** On the MVTec AD test split, replacing a defect block lowers the margin more than replacing a non-defect block (Table 7). Each replaced block is taken from the nearest normal training image of the product. Replacing all defect blocks flips 76.2% of the images detected at the default threshold, against 3.4% for as many random non-defect blocks (a comparator added at the results review).

As a map R4 is above chance but below the center prior at its own 8 × 8 resolution. It is evidence of dependence, not a good map. By the pre-registered rule, which reads the means, no other fill crosses chance. On a subset of 150 defective images, however, they attenuate R4 toward it (exploratory). Two of the three intervals include 0.5.

| Contrast | n | Mean [95% CI] |
|---|---|---|
| *Margin change, logits* | | |
| Single block: defect − non-defect | 879 | 0.201 [0.160, 0.247] |
| Single block, on-object blocks only | 494 | 0.184 [0.149, 0.219] |
| Same position on normal images, objects | 491 | 0.168 [0.132, 0.204] |
| Same position on normal images, textures | 267 | 0.566 [0.459, 0.686] |
| All at once: defect − seam-matched region, objects | 435 | 1.306 [1.193, 1.420] |
| Pasted: defect − normal blocks, objects | 495 | 1.360 [1.244, 1.475] |
| Pasted: defect − normal blocks, textures | 264 | 3.438 [3.263, 3.610] |
| *R4 as a map, per-image pixel AUROC* | | |
| Main fill | 879 | 0.630 [0.606, 0.654] |
| Main fill − center prior at 8 × 8 | 879 | −0.130 [−0.157, −0.102] |
| Main fill, subset | 150 | 0.632 [0.580, 0.683] |
| Blurred block, subset | 150 | 0.524 [0.468, 0.579] |
| Mean of normal images, subset | 150 | 0.568 [0.515, 0.620] |
| Inpainting, subset | 150 | 0.514 [0.462, 0.564] |

Table 7. Dependence of the decision on the defect region, MVTec AD test split. Single-block and all-at-once rows are larger drops of the margin; pasted rows are larger rises. The subset rows are exploratory; n counts images.

Location-matched controls, pre-registered after the first results and analyzed as a separate family, keep the effect, both restricted to on-object blocks and against the drop at the same position on normal images (Table 7).

A second amendment, written after the first results but pre-registered before its passes ran, adds content controls. Replacing all defect blocks at once lowers the margin more than replacing an equal, seam-matched region elsewhere on the object. Pasting the defect blocks into a matched normal image raises it more than pasting normal blocks of another image of the same product at the same places (for textures, pasted-normal test only). This also holds where the pasted normal content is at least as different in pixels.

Across products the single-block contrast above is not shown. Its equal-category mean is 0.172 with a t interval of [−0.025, 0.369]. Per product (Table 4) the two labeled below are cable and grid. Leather has the largest effect (without it, descriptively, [−0.022, 0.213]). These values come from the first follow-up and its amendment. The all-at-once and pasted contrasts were not analyzed this way. The statements here are therefore about these images, not across products.

Under the single-block location control the position-matched contrast is below zero for cable and transistor. This is descriptive and without foreground matching: our label-free foreground method fails its validity rule on them, so they lie outside the pooled object values above. Transistor's interval includes zero. The contrast is small for bottle, tile and pill. All at once, bottle and tile show large effects and pill stays small. No control excludes a response to foreign patterns that are not defects.

The head's attention is not faithful to this dependence. Its per-image rank correlation with the drops is −0.016 [−0.026, −0.007], inside the pre-registered negligible band of ±0.1. The correlation of the backbone's attention (8-layer mean), which agrees with the masks, is above zero but small and also inside ±0.1 (a secondary analysis with no pre-registered rule, 0.030 [0.020, 0.040]). Pointing at the defect does not by this test show that the map reflects what the decision depends on.

**A second release.** Clef 27B, post-trained from Qwen3.8-27B with the same head code [3], was run once on a rented GPU. Both releases detect without giving a useful map of the defect (Table 8). Clef-flash's routing attention ranks the defect low (per-image pixel AUROC). Clef 27B's is above chance but weaker than a fixed center prior. On MVTec AD it does no better than the shuffle and, within the object, it still ranks the defect low (a secondary analysis). On VisA it beats the shuffle by little and lies slightly above chance within the object (a secondary analysis). The comparison cannot separate model size from backbone generation or training recipe.

Clef 27B was not shown to detect better. On MVTec AD the interval of the difference with Clef-flash excludes zero but crosses the margin (inconclusive), mostly from two categories. On VisA the two were equivalent (Table 8). On MVTec AD 2 the primary comparison was inconclusive, while across all lighting conditions Clef 27B was worse (−0.094 [−0.159, −0.030]).

| Clef 27B | MVTec AD | VisA |
|---|---|---|
| R1 | 0.549 [0.539, 0.559] | 0.589 [0.577, 0.601] |
| R1 − center prior | −0.211 [−0.225, −0.197] | −0.236 [−0.249, −0.222] |
| R1 − shuffle | −0.004 [−0.017, 0.009] | 0.023 [0.009, 0.038] |
| R1 within the object | 0.458 [0.444, 0.474] | 0.519 [0.505, 0.533] |
| Image AUROC − Clef-flash | −0.030 [−0.049, −0.013] | −0.012 [−0.028, 0.005] |

Table 8. Clef 27B's routing attention (per-image pixel AUROC and paired differences on defective images) and detection against Clef-flash (means or differences, 95% CIs). The within-object row is a secondary analysis on the categories with a valid foreground.

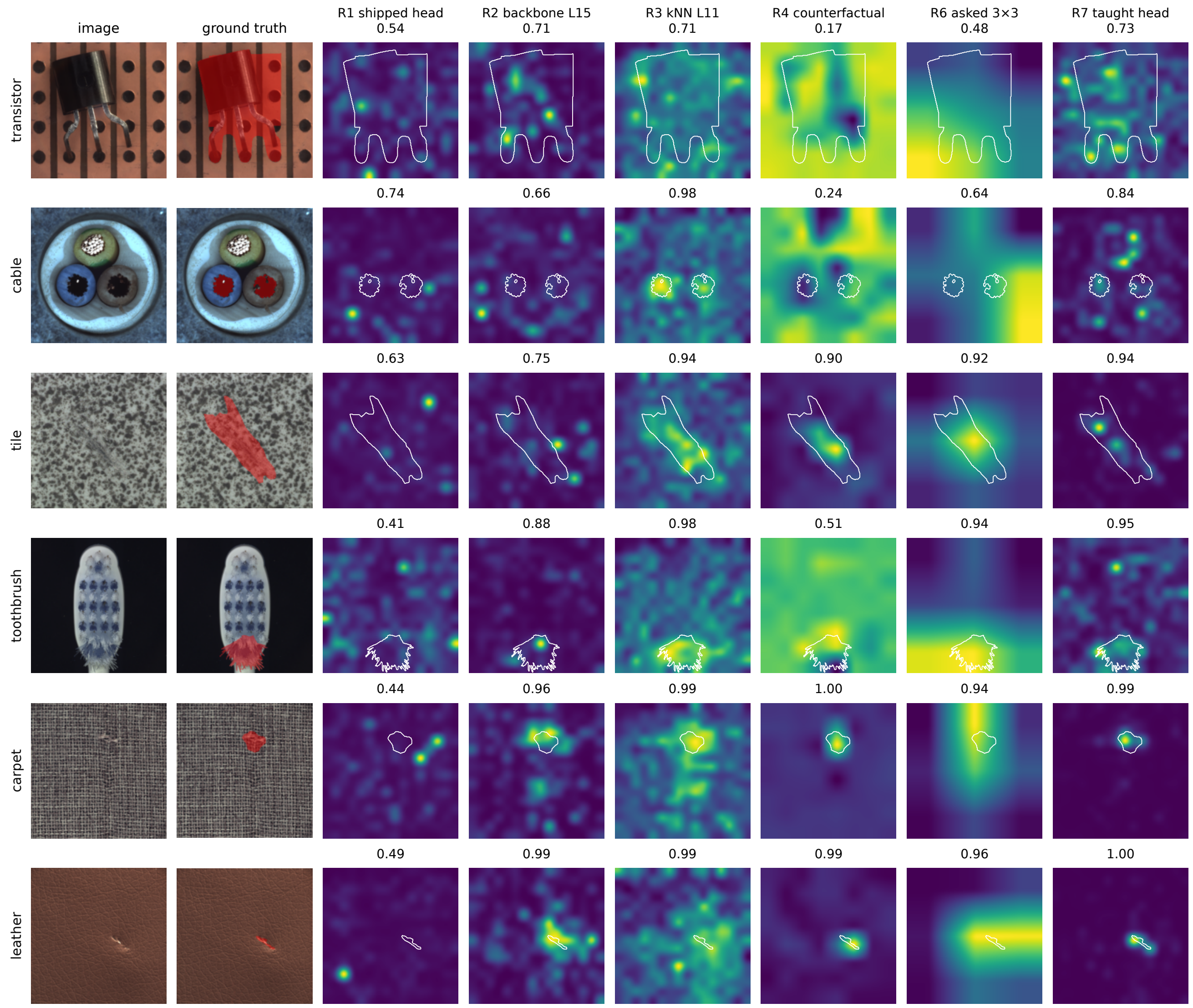

Figure 5. Rule-selected MVTec AD test images (two lowest, two middle and two highest categories by the taught head's per-category AUROC; the lower-median image in each) with the mask and the maps R1, R2, R3, R4, R6 and R7. The rule favors the taught head's range, so these R1 tiles are not typical of the mean of R1.

Two pre-registered checks failed. The first asked that the rented machine reproduce single maps and probabilities (r $\geq 0.99$ and $|\Delta p| \leq 0.01$ per image, on a list of 70 images, the maps on those from MVTec AD and VisA). The largest $|\Delta p|$ was 0.024. The second was the probe's correlation with Clef-flash (r $>$ 0.5 on at least 80% of 27 images); none reached it, median 0.344. An amendment, made after these failures and before the full run, replaced the first with looser checks on the decision and a check on aggregates. It also stopped gating on the second, leaving the raster check to validate the mapping.

Over the full re-run, Clef-flash's head maps correlate 0.876 [0.871, 0.882] between the machines on MVTec AD and 0.898 [0.893, 0.902] on VisA. A single image's pixel AUROC moves by 0.018 on average on MVTec AD, while the mean R1 there differs by 0.000 [−0.002, 0.002]. Only aggregates are therefore reported for the larger release. Comparisons between releases use Clef-flash re-run on the rented machine, and all other Clef-flash numbers come from our own machine.

## 7. Recovering localization

Table 12 compares three routes on the MVTec AD test split.

**Asking** adds one yes/no question per cell of a 3 × 3 grid to the same request; the phrasing (region words) was chosen on the dev split. The map is above chance, the center prior (by 0.065 [0.048, 0.083]) and the head's attention, but not beyond the foreground prior (−0.010 [−0.032, 0.011]).

| Run | Per-image pixel AUROC | Detection change |
|---|---|---|
| Seed 0 | 0.896 [0.884, 0.907] | −0.042 [−0.064, −0.021] |
| Seed 1 | 0.888 [0.876, 0.900] | −0.031 [−0.052, −0.011] |
| Seed 2 | 0.895 [0.883, 0.906] | −0.043 [−0.064, −0.023] |
| No alignment term | 0.619 [0.605, 0.633] | −0.012 [−0.027, 0.001] |

Table 9. The head taught on VisA, MVTec AD test split: its attention, and its image AUROC minus the shipped head's. Seeds use the selected loss weight. Intervals are 95%, over images only.

The grid rule counts a hit when the top cell overlaps the mask, and it excludes images whose mask covers at least 7 of the 9 cells. Under this rule, the hit rate of its top cell is 0.784 [0.756, 0.812] on 818 images. Always naming the center gives 0.603 [0.568, 0.636], and the chance level is 0.285. Detection stays equivalent (0.001, [−0.002, 0.005]). At 8 × 8 the map is level with the center prior, and on VisA it is below the center prior.

**Teaching** retrains the head alone on cached states (the backbone is not touched). It uses CutPaste-style synthetic defects in VisA's normal images [26] and a loss that aligns the attention to the synthetic masks. On MVTec AD, never seen during adaptation, the taught attention beats every control including the foreground prior. This holds in all three seeds at the selected loss weight (Table 9; Table 12 gives the first). Without the alignment term (one run), the attention moves toward the defect but not beyond the center prior.

Detection decreased. In each seed the taught head's image AUROC minus the shipped head's has an interval that crosses the ±0.03 margin, so the change is inconclusive against it (Table 9). Accuracy on BoolQ [27] shows no detectable change (89.8% against 89.7%). In the reverse direction (MVTec AD to VisA), at the selected loss weight, the attention does not beat the center prior and BoolQ accuracy collapsed in two of three runs. Two of the three seeds in that direction were added in an amendment written after the first results.

The taught map is plausible but shows no faithfulness beyond ±0.1. We re-ran the block replacement with the taught head (also an amendment after first results). On non-defect blocks, where the defect blocks themselves are left out, its attention's rank correlation with the taught decision's drops is given in Table 10. The taught decision depends on the defect region as a whole about as much as the shipped one's (no paired test). Its single-block contrast between defect and non-defect blocks, however, is not shown (−0.054 [−0.127, 0.024]).

**Reading** the backbone needs no change to the model or the request. The probe at the dev-selected layer 11,

| Map | Blocks | Rank correlation |
|---|---|---|
| R1, shipped head | all | −0.016 [−0.026, −0.007] |
| R2, 8-layer mean | all | 0.030 [0.020, 0.040] |
| R7, taught head | all | 0.002 [−0.011, 0.013] |
| R7, taught head | non-defect | 0.000 [−0.010, 0.010] |
| R3, layer 11 | non-defect | −0.029 [−0.040, −0.018] |
| R3, final states | non-defect | −0.028 [−0.038, −0.017] |

Table 10. Per-image rank correlation of each map with the decision's drops under block replacement, MVTec AD test split (means, 95% CIs). R7 is compared with the taught decision, the others with the shipped one. The all-blocks rows for R2 and R7 are secondary analyses, the R7 rows come from an amendment after first results and the R3 rows from a post-results follow-up.

with sixty normal images, gives the best tested map among the Clef-flash readouts (the taught head minus the probe is −0.037 [−0.046, −0.029]). It remains an information test, not a readout of the decision. On non-defect blocks, tracking of the decision's drops is not shown (Table 10; a post-results follow-up). The backbone's attention at its dev-selected layer 15 needs no normal images, under the inspection request. It scores about as well as the taught head, a comparison not tested.

None of the routes matches the few-shot localizers on the same images, and against zero-shot WinCLIP they are mixed. On the test split, descriptively (no test), our zero-shot WinCLIP map scores about the backbone's attention, above asking and below the taught head and the probe. WinCLIP+ with one normal image and PatchCore with sixteen score above every Clef-flash readout (Table 12). Table 13 gives the full-set values of the maps fixed in advance. Figure 5 shows rule-selected examples.

**Cost.** Table 11 gives the median time per image on our hardware. Reading and teaching add no measurable time to the decision. The probe needs a one-off bank of normal images. Asking at 3 × 3 was timed in that experiment's own run. The block-replacement map R4 is a tool for analysis.

On the same machine, Clef-flash runs with part of its backbone offloaded, and the two baselines run without offloading and unoptimized. Teaching adds one head training. Among the Clef-flash readouts, differences of a few hundredths of a second are within the image-to-image spread (10th–90th percentile).

| Method | One-off need | s per image |
|---|---|---|
| R0, the decision | — | 0.609 |
| R2, backbone attention | — | 0.595 |
| R7, taught head | one head training | 0.610 |
| R3, probe of the final states | bank of 60 images | 0.625 |
| R6, 3 × 3 | — | 0.767 |
| R4, 8 × 8 | bank of 60 images | about 40 |
| WinCLIP, zero-shot, $240^2$ input | — | 0.044 |
| PatchCore, $256^2$ input | bank of 16 images | 0.010 |

Table 11. Median seconds per image on one machine.

| Route | Readout | Needs | Per-image pixel AUROC |
|---|---|---|---|
| Asking | R6, 3 × 3 | 9 more questions | 0.816 [0.805, 0.827] |
| Teaching | R7 | synthetic defects, head training | 0.896 [0.884, 0.907] |
| Reading | R2, layer 15 | the inspection request, labeled layer choice | 0.888 [0.879, 0.898] |
| Reading | R3, layer 11 | 60 normal images, labeled layer choice | 0.933 [0.926, 0.939] |
| Shipped | R1 | nothing | 0.391 [0.379, 0.404] |
| Reference | WinCLIP zero-shot | nothing | 0.877 [0.866, 0.886] |
| Reference | WinCLIP+ | 1 normal image | 0.964 |
| Reference | PatchCore | 16 normal images | 0.979 |

Table 12. Routes to a map, MVTec AD test split (dev-selected layers and the taught head on the test split only). The taught map's rank correlation with the decision's drops stays inside ±0.1, and with the taught head detection decreased (inconclusive against ±0.03). Reference rows are our runs of the baselines, descriptive, with no test; WinCLIP+ and PatchCore are averaged over banks.

| Map | Normal images | Per-image | Pooled | AUPRO |
|---|---|---|---|---|
| R1, head attention | 0 | 0.391 | 0.460 | 0.093 |
| R2, backbone attention (8-layer mean) | 0 | 0.863 | 0.781 | 0.661 |
| R3, probe of the final states | 60 | 0.897 | 0.871 | 0.719 |
| WinCLIP, zero-shot (our run) | 0 | 0.875 | 0.823 | 0.618 |
| WinCLIP+ (our run) | 1 | 0.963 | 0.938 | 0.841 |
| PatchCore (our run) | 16 | 0.978 | 0.971 | 0.900 |

Table 13. Clef-flash maps fixed in advance against zero-shot and few-shot localizers, full MVTec AD test set, same masks and post-processing (per-image and pooled pixel AUROC, AUPRO to 30% FPR; WinCLIP+ and PatchCore averaged over five banks). Descriptive; our zero-shot WinCLIP run does not reproduce its published numbers.

# 8. Discussion and limitations

The dissociation is consistent with a decision that passes through a few summary vectors: the head need not attend to the defect to decide, and nothing in its post-training, as the vendor describes it, asks it to [3]. Why it ranks the defect below chance, rather than at chance, remains open. For integrators and explanation tools using these two releases, the lesson is that the head's recomputed attention is not a useful map of the defect; whether this holds for decision heads in general is a hypothesis. A location should come from a tested route, within the limits of Sec. 7. Which part of the inspection request makes the backbone's attention beat a center prior is left for a study that varies the field name, task, product and note one at a time. R2 moves with the request while R1 hardly does, which fits a head whose query barely changes, but the test cannot say why.

| Part of the dissociation | MVTec AD | VisA |
|---|---|---|
| head attention ranks the defect low, image-specifically (per-image pixel AUROC) | on average; 8 below, 1 above, 3 null, 3 inconclusive of 15 | on average; no per-category test |
| object preferred, lower within it | yes | yes |
| hiding image positions from the head's attention leaves detection unchanged | yes | not run |
| backbone attention points at the defect, under the inspection request | yes | not run |
| states carry the location (probe) | yes | partly† |
| decision depends on the defect region | test-split images; not shown across products | not run |
| no useful head map in either release | yes | yes |

Table 14. Which dataset supports each part of the dissociation. Rows concern Clef-flash unless they name a release. †Within the object (exploratory); marginal on the whole image. The central claim rests on MVTec AD.

The limits are those of a case study: one model family documented only by its vendor, with a confounded comparison between releases, and two follow-ups written and pre-registered after a review of the first results. The central claim rests on MVTec AD; several parts were not run on VisA (Table 14). Public benchmarks may be in the backbone's training data; only MVTec AD has dedicated symptom tests. The dependence test uses one main fill, on which the strength of the map depends, and one grid, and does not control for foreign patterns that are not defects. The reading layers were chosen on labeled dev images. Timings come from one machine with CPU offload. Our zero-shot WinCLIP runs on MVTec AD and VisA do not reproduce published numbers.

# 9. Conclusion

A released decision model, Clef-flash, detects industrial anomalies without seeing images of the product, and unevenly across benchmarks and products (Sec. 5). Its head does not point at the anomalies it detects. On MVTec AD the head's attention ranks the defect low, by per-image pixel AUROC: below 0.5 on average and, by an adjusted per-category test, below 0.5 in 8 of 15 categories, above it in 1, null in 3 and inconclusive in 3 (post-results follow-up). The same follow-up finds it below 0.5 both on the images the decision detects and on those it misses, with no difference between the two shown. A larger release does not give a useful head map either.

The information is present elsewhere in the model. Under the inspection request the backbone's attention points at the defect; with another question it falls below the center prior, and with the inspection context removed as well, below chance (post-results follow-up). The backbone's states carry the location when probed with normal images. On the test-split images the decision depends on the defect region, though the single-block effect is not shown across products and a response to foreign patterns is not excluded.

A location should therefore not be read from the shipped heads' attention. It can come from a tested route, within limits, and the three routes we tested do so unequally. Reading the backbone gives the best Clef-flash map on MVTec AD: a probe of its states with sixty normal images, which is an information test rather than a readout of the decision. The backbone's attention needs no normal images, but its score depends on the inspection request and it scores lower; both are read at layers chosen on labeled images. Asking region questions needs no normal image, but does not go beyond a foreground prior on objects, is below the center prior on VisA and, descriptively on the MVTec AD test split, scored below our zero-shot WinCLIP run. Teaching the head beats every control on MVTec AD, but detection decreased (inconclusive against the $\pm 0.03$ margin), the taught map's correlation with the decision's drops stays inside $\pm 0.1$, and in the reverse direction it does not beat the center prior. Where normal images of the product are available, dedicated few-shot localizers scored higher than all three routes, descriptively, on MVTec AD.